\documentclass[sn-mathphys,iicol]{sn-jnl}
\usepackage{graphicx}
\usepackage{subfigure}
\makeatletter
\renewcommand{\@thesubfigure}{\hskip\subfiglabelskip}
\makeatother
\usepackage{picins}
\usepackage{xcolor}
\usepackage{lmodern}
\usepackage{amsmath}

\jyear{2022}%

\theoremstyle{thmstyleone}
\theoremstyle{thmstyletwo}%

\theoremstyle{thmstylethree}%

\begin{document}

\title[CF-YOLO]{CF-YOLO: Context-Aware Feature Refinement for Camouflaged Industrial Micro-Defect Detection}

\author[1]{\fnm{Xinda} \sur{Yu}}\authordetails{Xinda Yu, E-mail:} \email{2025388032@stu.zjhu.edu.cn}

\author[1]{\fnm{Kunxin} \sur{Zheng}}\authordetails{Kunxin Zheng, E-mail:} \email{2024092411@stu.zjhu.edu.cn}

\author[2]{\fnm{Chunan} \sur{Yu}}\authordetails{Chunan Yu, E-mail:} \email{yuchun\_an@njust.edu.cn}

\author[1]{\fnm{Qingbo} \sur{Song}}\authordetails{Qingbo Song, E-mail:} \email{2025388020@stu.zjhu.edu.cn}

\author[1]{\fnm{Hao} \sur{Xiao}}\authordetails{Hao Xiao, E-mail:} \email{xiaohao@zjhu.edu.cn}

\author[1]{\fnm{Ying} \sur{Zang}}\authordetails{Ying Zang, E-mail:} \email{02750@zjhu.edu.cn}

\author*[1]{\fnm{Jie} \sur{Liu}}\authordetails{Jie Liu, E-mail:} \email{liujie52@zjhu.edu.cn}

\affil[1]{\orgdiv{School of Information Engineering}, \orgname{Huzhou University},\orgaddress{ \city{Huzhou}, \postcode{313000}, \country{China}}}

\affil[2]{\orgdiv{School of Computer Science and Engineering}, \orgname{Nanjing University of Science and Technology},\orgaddress{ \city{Nanjing}, \postcode{210000}, \country{China}}}

\abstract{Automated detection of surface micro-defects on industrial components, such as copper tubes, is critically important for quality assurance but remains challenging due to the minute scale of anomalies and their visual camouflage against complex backgrounds. These factors lead to weak feature representations and high rates of false positives and missed detections. To address these issues, we propose a novel real-time detection framework designed for efficient context perception and feature refinement. Our method integrates a Context-Perception Aggregation Module (CPAM), which synergises large-kernel perception for macro-texture context and small-kernel aggregation for sharp boundary delineation, effectively breaking the background camouflage. Furthermore, a Feature Additive Refinement Module (FARM) employs a linear-complexity additive token mixer to globally verify and refine the representation of fine-grained anomalies, suppressing noise-induced errors. To support research in this domain, we introduce the Copper Tube Defect Dataset (CTDD),  {a manually annotated benchmark containing 1,847 images and 4,898 bounding-box defect instances from copper-tube inspection scenarios}. Extensive experiments demonstrate that our detector achieves strong and consistent performance on CTDD, outperforming representative baseline detectors, including YOLOv11, by 2.2\% in mAP@50 and 3.9\% in Precision while maintaining real-time inference speed. This work provides a robust and efficient solution for high-precision industrial inspection, bridging the gap between contextual understanding and detailed feature analysis. Our code and model are available at: https://github.com/Yu-Xinda/CF-YOLO-Context-Aware-Feature-Refinement-for-Camouflaged-Industrial-Micro-Defect-Detection}

\maketitle
\twocolumn[\section{Introduction}]
    In the era of Industry 4.0, intelligent manufacturing has become a cornerstone for enhancing production efficiency and product quality~\cite{app8091575,Xiang2025,Song2025,Shi2025,CHENG2026152}. 
    As a critical component in refrigeration, plumbing, and heat exchange systems, the surface integrity of copper tubes directly dictates the sealing performance and operational safety of the final equipment. Consequently, the transition from labor-intensive manual inspection to automated defect detection is not merely an option but an imperative necessity for the industry~\cite{DBLP:journals/ijcv/BergmannBFSS21}. However, despite rapid advancements in computer vision, establishing a high-precision, real-time detection system for industrial micro-defects remains a formidable task, presenting complexities that far exceed those encountered in generic object detection scenarios~\cite{DBLP:journals/jim/TabernikSSS20,DBLP:journals/iswa/NikoueiBNTASM25,LI2026119556}.
   
\begin{figure}[h]
\centering
\includegraphics[width=0.42\textwidth]{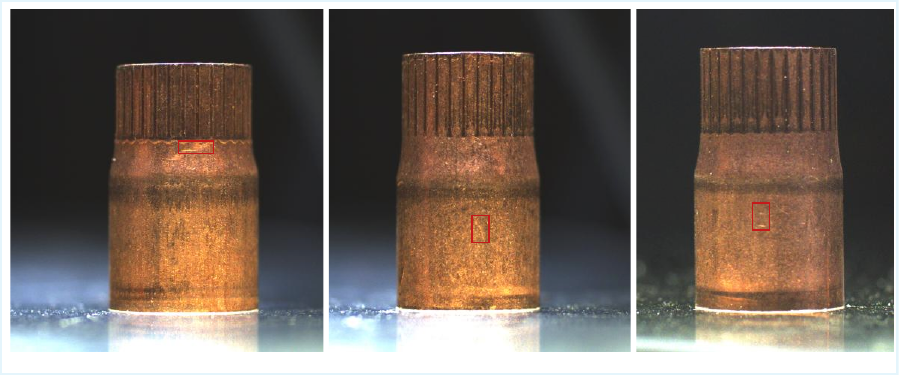}
\caption{Visualization of challenging defects with severe background camouflage. The anomalies are nearly indistinguishable due to highly similar textures with the background. Their locations are indicated by red bounding boxes.} 
\label{fig:defect_part}
\end{figure}

    The primary obstacles hindering effective detection in this domain are twofold: micro-scale targets and background camouflage, as shown in Fig.~\ref{fig:defect_part}. First, the defects typically appear as faint, fine-grained scratches that occupy a negligible proportion of image pixels, leading to weak feature representations that are easily lost during network downsampling~\cite{DBLP:conf/cvpr/LinDGHHB17,10019311}. Secondly, and more critically, these defects often exhibit visual patterns—such as color and texture—that are highly homologous to the copper substrate. This ``camouflaged" nature makes it exceptionally difficult to distinguish genuine anomalies from complex background textures or surface oxidation noise, resulting in a high rate of both missed detections and false positives~\cite{DBLP:conf/iccv/ZhangLL0JX23,Zavrtanik_2021_ICCV}.{Moreover, environmental variation and complex region-dependent textures further increase feature ambiguity in industrial inspection scenes~\cite{CEAL,MSEmbGAN}.}

\label{sec1}

    To address these issues, various effective deep learning paradigms and advanced model architectures have been proposed~\cite{EAPT,UTMCR,Adaptive,MNGNAS}. Specifically,mainstream approaches typically employ Convolutional Neural Networks (CNNs) or Vision Transformers (ViTs)~\cite{DBLP:conf/iclr/DosovitskiyB0WZ21}.  {Recent detectors such as the YOLO series} utilize sophisticated blocks for efficiency, relying on the stacking of compact convolutional layers to extract fine-grained patterns. These mechanisms suffer from a restricted receptive field, limiting their ability to perceive the broader contextual consistency required to identify camouflaged defects~\cite{DBLP:conf/nips/LuoLUZ16}. Conversely, while ViTs offer global context via self-attention\cite{DBLP:conf/iclr/DosovitskiyB0WZ21}, their quadratic computational complexity renders them unsuitable for real-time industrial deployment~\cite{DBLP:journals/pami/00020C0GLTXXXYZ23}. Furthermore, the attention mechanism tends to smooth out high-frequency details, which is detrimental to preserving the sharp edges of micro-defects~\cite{DBLP:conf/iclr/ParkK22}.  {Recent studies have shown that structure-aware feature interaction and fine-grained representation learning are crucial for capturing subtle visual discrepancies in complex scenes~\cite{SATNet,ContrastiveDecoupling}.}Thus, a solution that balances global context perception, local detail preservation, and inference speed is urgently needed~\cite{DBLP:conf/nips/LiYWHETWR22}.
    
    In this paper, we propose CF-YOLO, which stands for Context-perception and Feature-refinement YOLO. Additionally, the acronym ``CF" implicitly embodies the ``Coarse-to-Fine" detection philosophy integrated into our architecture, which synergizes the strengths of large-kernel convolutions and additive attention. First, to tackle the ``background camouflage" issue, we introduce the Context-Perception Aggregation Module (CPAM). By decoupling feature processing into large-kernel contextual perception and small-kernel dynamic aggregation, CPAM effectively models the macro-texture of the copper surface, enabling the network to differentiate anomalies from the background. Second, to address the ``micro-scale" issues, we design the Feature Additive Refinement Module (FARM) following the spatial attention stage. Adopting a linear-complexity additive token mixer, FARM integrates spatial details with channel-wise semantic attributes. This allows for global semantic verification of candidate features, effectively refining representations to distinguish true defects from intricate surface noise.
    
    Beyond algorithmic improvements, the lack of high-quality, publicly available datasets for copper tube defects significantly hinders research progress in this community. To bridge this gap, we construct and release a comprehensive dataset named the Copper Tube Defect Dataset (CTDD).  {CTDD contains 1,847 copper-tube images and 4,898 single-class bounding-box defect annotations collected under industrial lighting conditions. Recent studies have highlighted the importance of transparent datasets and reproducible evaluation in visual-computing research~\cite{dataset_distillation,retinal_dkd_lancet_2025}. CTDD therefore provides a standardized benchmark for industrial defect detection.} The release of CTDD not only serves as a rigorous benchmark for evaluating our proposed method but also provides a valuable resource for data augmentation and future research in the field of industrial surface inspection. The main contributions of this work are summarized as follows:

\begin{itemize}
\item  {We propose a real-time detector tailored for industrial micro-defect detection, achieving an explicit accuracy-efficiency trade-off.}
\item  {We design two task-adapted modules: CPAM for context-guided local aggregation and FARM for high-level feature refinement.}
\item  {We contribute CTDD, an annotated copper tube defect detection benchmark with 1,847 images and 4,898 defect instances.}
\end{itemize}

\section{Related work}\label{sec2}
\subsection{Object Detection Methods}
    Object detection, a fundamental vision task, involves localizing and classifying instances. Driven by deep learning, recent methodologies are extensively adopted for high-precision tasks like industrial defect detection.
    
    Two-stage detectors like Faster R-CNN~\cite{DBLP:conf/nips/RenHGS15} are renowned for high accuracy but suffer from slower inference compared to one-stage counterparts. Building on this, Cascade R-CNN~\cite{DBLP:conf/cvpr/CaiV18} employs a multi-stage structure to enhance robustness, though this inevitably increases training time and computational demands.
    
    Conversely, one-stage detectors prioritize speed. SSD~\cite{DBLP:conf/eccv/LiuAESRFB16} utilizes multi-scale feature maps but exhibits limitations in managing class imbalance. RetinaNet~\cite{DBLP:conf/iccv/LinGGHD17} addresses this via Focal Loss, effectively mitigating sample imbalance, albeit sometimes yielding lower overall accuracy. 
    
    Recently, the YOLO series has evolved rapidly to address complex industrial constraints. For instance, Yolov8-HAC~\cite{Yolov8-HAC} introduced a safety helmet detection model optimized for complex underground coal mine scenes, validating the effectiveness of YOLO in harsh environments. Similarly, YGC-SLAM~\cite{YGC-SLAM} demonstrated that improving YOLOv5 with geometric constraints significantly enhances robustness in dynamic indoor environments.  {Industrial variants such as AFF-YOLO~\cite{aff_yolo} and YOLO-RFF~\cite{yolo_rff} further improve defect detection through enhanced feature representation and multi-scale modeling, demonstrating the value of task-oriented architectural adaptation.}

\subsection{Surface Defect Detection}
    Surface defect detection plays a critical role in industrial production, prompting extensive research into automated solutions. Traditional methods typically rely on hand-crafted features and statistical learning. Common techniques range from edge detection operators \cite{DBLP:journals/tsmc/Otsu79}and Otsu-based thresholding to Fourier image reconstruction\cite{DBLP:journals/prl/Ng06}. In specific domains like TFT-LCD and mobile screen, low-rank matrix models~\cite{DBLP:journals/csur/ZhouYZY14} and contour-based registration have been utilized to identify defects. However, the heavy reliance on manual feature design often renders these traditional methods less effective in complex environments or when facing intricate defect patterns.
    
    Recently, the paradigm has shifted significantly towards deep learning. 
    To preserve texture and fine-grained details, architectures such as SDDNet~\cite{DBLP:conf/mm/CongGCZZK23} integrate feature retaining blocks and densely connected modules. Addressing the challenge of small object detection, particularly in printed circuit boards (PCBs)\cite{DBLP:journals/pami/ChenPKMY18}, novel feature fusion methods like the atrous spatial pyramid pooling-balanced FPN\cite{DBLP:conf/cvpr/LinDGHHB17} have been introduced. 
    
    Advanced models like ETDNet~\cite{9353254} incorporate lightweight vision transformers and decoupled heads to optimize both classification and regression.  {In parallel, efficient context modeling has attracted increasing attention. Methods such as LSKA~\cite{lska} leverage large-kernel attention to capture long-range dependencies, while SwiftFormer~\cite{swiftformer} employs lightweight additive attention to improve feature interaction with low computational cost.} To tackle data scarcity, frameworks employing Generative Adversarial Networks (GANs)~\cite{DBLP:conf/nips/GoodfellowPMXWOCB14} and global context-based augmentation have been developed to boost performance via synthetic data generation.

\section{Proposed Method}
\label{sec3}

\begin{figure*}[h]
\centering
\includegraphics[width=0.8\textwidth]{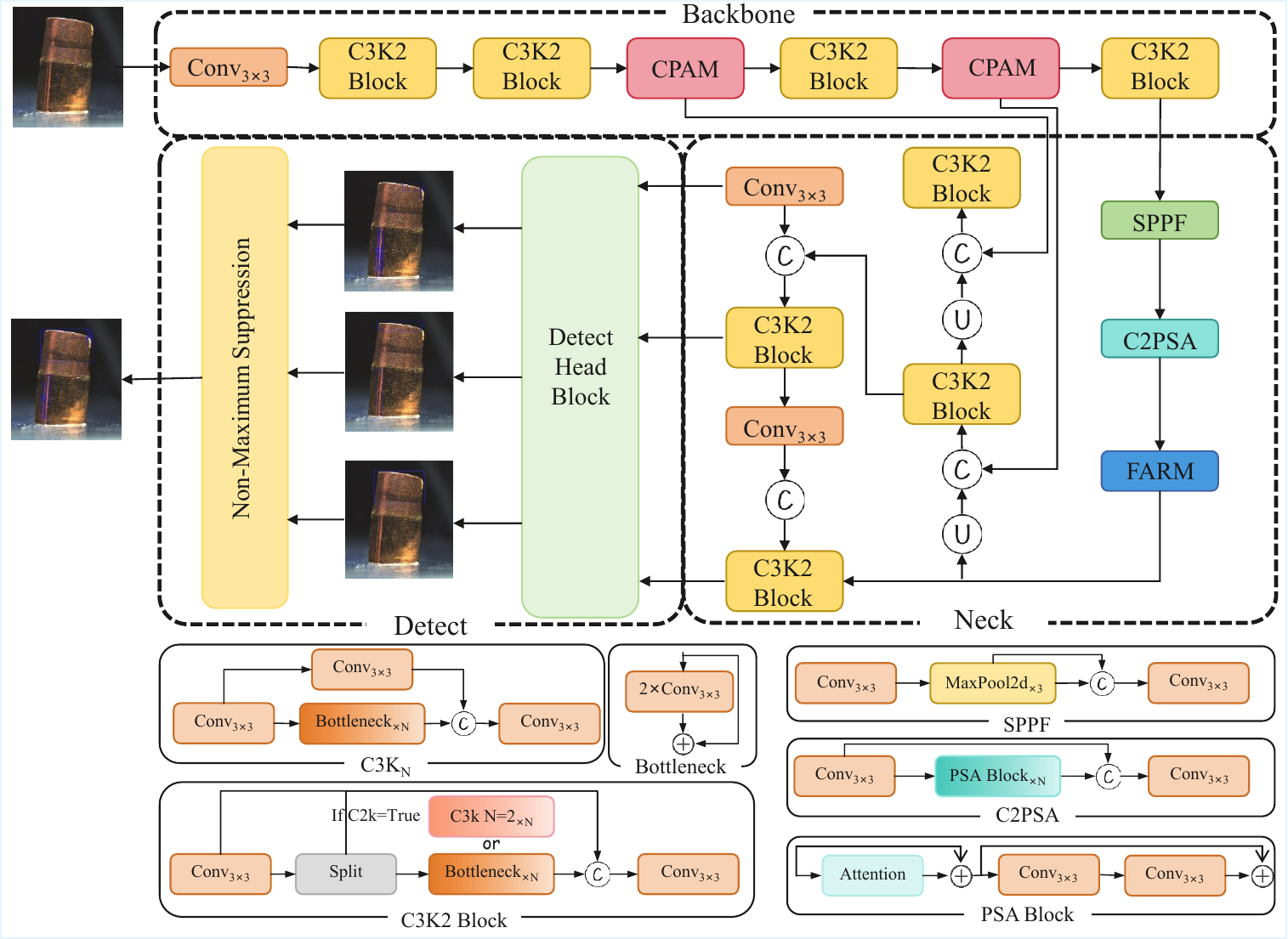}
\caption{Schematic illustration of CF-YOLO for industrial copper tube defect detection, integrating CPAM for context perception and FARM for semantic refinement.} 
\label{CFYOLO}
\end{figure*}

\subsection{Preliminaries and Problem Formulation}
    In this work, we formulate the micro-defect detection task as a supervised dense prediction problem aimed at localizing and classifying anomalies on industrial surfaces. Let the dataset be denoted as $\mathcal{D} = \{(I_i, \mathcal{Y}_i)\}_{i=1}^{N}$, where $I_i \in \mathbb{R}^{H \times W \times 3}$ represents an input image and $\mathcal{Y}_i$ denotes the corresponding ground-truth set. Each ground-truth object is parameterized by a bounding box $ b = (cx, cy, w, h)$ and a class label $c \in \{1, \dots, N\}$, where $N$ denotes the number of detecting classes. Our primary objective is to construct a deep neural network $\mathcal{F}$ that learns a mapping function from the input image $I_i$ to a set of predictive distributions. The feature extraction process transforms the input image into a multi-scale feature pyramid $\mathcal{P} = \{P_3, P_4, P_5\}$, where $P_l \in \mathbb{R}^{\frac{H}{2^l} \times \frac{W}{2^l} \times C_l}$ represents the features at a stride of $2^l$, which serves as the downsampling ratio determining the spatial resolution of each level.

\subsection{Overall Architecture of CF-YOLO}
    The proposed \textbf{CF-YOLO} architecture is designed to balance macro-contextual perception with micro-semantic refinement, addressing the specific challenges of background camouflage and weak feature representation in industrial scenarios. As illustrated in Fig.~\ref{CFYOLO}, the network comprises three hierarchical components: a Context-Aware Backbone, a Feature-Refining Neck, and a Decoupled Head. While constructed upon the YOLOv11 framework, CF-YOLO redefines the feature encoding pipeline through two strategic enhancements. 

    Given an input image $I_i \in \mathbb{R}^{H \times W \times C}$, the backbone initially utilizes $Conv_{3\times3}$ stacked C3k2 blocks to extract foundational feature representations, denoted as $F_1$ and $F_2$. These features are subsequently processed by CPAM modules to effectively disentangle camouflaged anomalies from the background. By alternating these C3k2 blocks and CPAM modules, the network constructs a multi-scale feature pyramid $\mathcal{F} = \{F_3, F_4, F_5\}$.  {The selected stride-8, stride-16, and stride-32 pyramid balances micro-defect sensitivity and contextual reasoning: the stride-8 feature preserves fine spatial cues for small low-contrast defects, while stride-16 and stride-32 features provide broader context for suppressing oxidation, machining texture, and other background-like responses. Using lower-stride features would increase memory and latency, whereas removing $F_3$ would risk losing weak defect evidence during downsampling.} Prior to the multi-scale fusion stage, the fine-resolution feature map $F_5$ undergoes specific refinement via the FARM module to explicitly capture micro-scale defects. Following this enhancement, the pyramid features interact through a cross-scale fusion neck to yield the refined set $\mathcal{P} = \{P_3, P_4, P_5\}$. Finally, these features are fed into prediction heads to generate multi-scale outputs $\mathcal{Y}_l = (\mathbf{O}_{cls}^l, \mathbf{O}_{reg}^l)$, which are consolidated via NMS to produce the final detection $Y_{final}$. 
    

\subsection{Context-Perception Aggregation Module}
    Motivated by the feature extraction limitations stemming from the limited receptive field in native YOLO architectures, we introduce the Context-Perception Aggregation Module (CPAM). Let $\mathbf{F}_{in} \in \mathbb{R}^{H_i \times W_i \times C_{in}}$ represents the feature processed from the C3K2 block. CPAM decouples the feature processing into two sequential stages: Large-Kernel Perception (LKP) and Small-Kernel Aggregation (SKA).

\begin{figure}[h]
\centering
\includegraphics[width=0.3\textwidth]{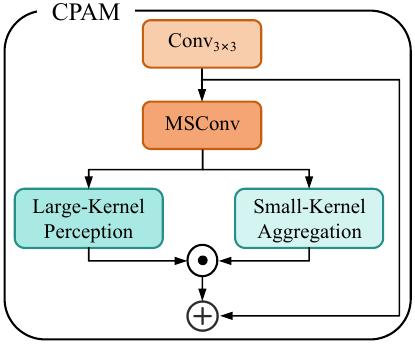}
\caption{The architecture of the Context-Perception Aggregation Module (CPAM), illustrating the decoupling of Large-Kernel Perception (LKP) and Small-Kernel Aggregation (SKA).} 
\label{CPAM}
\end{figure}

    The LKP stage aims to capture the macro-texture of the background with minimal computational overhead. We first reduce the channel dimension to $C' = C/2$ via a point-wise convolution ($\text{PW}$) to maintain efficiency. Subsequently, a large-kernel depth-wise convolution ($\text{DW}$) with a kernel size of $K_L \times K_L$ is applied to perceive a wide contextual range $\Omega_{L}$. A final $\text{PW}$ convolution projects the features to generate a dynamic weight tensor $\mathbf{W}_{agg} \in \mathbb{R}^{H \times W \times D}$, which encodes the spatial context guidance. This operation is formulated as:
\begin{equation}
\mathbf{W}_{agg} = \text{PW}_2(\text{DW}_{K_L}(\text{PW}_1(\mathbf{F}_{in})))
\end{equation}

    Following perception, the SKA stage performs precise feature fusion guided by $\mathbf{W}_{agg}$. The weight tensor is reshaped to form dynamic kernels of size $K_S \times K_S$. We partition the input $\mathbf{F}_{in}$ into $G$ groups to facilitate group-wise dynamic convolution. The final output $\mathbf{F}_{cp}$ is obtained by aggregating the local neighborhood $\mathcal{N}_{K_S}$ using the generated weights:
\begin{equation}
\mathbf{F}_{cp} = \mathcal{A}_{dynamic}(\mathbf{F}_{in}, \mathbf{W}_{agg}) = \mathbf{W}_{agg} \circledast \mathbf{F}_{in}
\end{equation}
where $\circledast$ denotes the group-wise dynamic convolution. By coupling large-field perception with local aggregation, CPAM effectively delineates defect boundaries from complex backgrounds.

\subsection{Feature Additive Refinement Module}
    Although alternating feature extraction using the C3K2 module and CPAM effectively partially alleviates the "target camouflage" problem to some extent and improves recognition accuracy, this architecture is still insufficient in capturing the features of small targets. To address this limitation, we introduce a Feature Additive Refinement Module (FARM) to enhance the network's ability to accurately perceive and capture small targets. FARM addresses this issue via a linear-complexity global verification mechanism. Let $F_{i} \in \mathbb{R}^{W^{'} \times H^{'} \times C}$ be the flattened input features. FARM projects $\mathbf{F_i}$ into Query ($\mathbf{Q}$), Key ($\mathbf{K}$), and Value ($\mathbf{V}$) spaces via linear transformations.

\begin{figure}[h]
\centering
\includegraphics[width=0.5\textwidth]{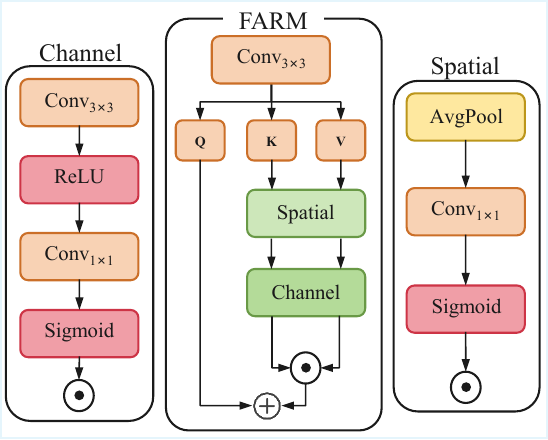}
\caption{The architecture of the Feature Additive Refinement Module (FARM). Adopting the Convolutional Additive Token Mixer (CATM) paradigm, it performs global semantic refinement via dual-domain (spatial and channel) interactions and linear-complexity additive similarity.} 
\label{FARM}
\end{figure}

    Unlike quadratic self-attention mechanisms, FARM employs a \textbf{Convolutional Additive Token Mixer}. We define a context mapping function $\Phi(\cdot)$ that integrates dual-domain interactions. The spatial interaction is modeled by a $3\times3$ depth-wise convolution followed by a Sigmoid activation $\sigma$, while the channel interaction is modeled by global average pooling ($\text{AvgPool}$) and channel-wise modulation. For a generic token tensor $\mathbf{T} \in \{\mathbf{Q}, \mathbf{K}\}$, the mapping is defined as:
\begin{equation}
\Phi(\mathbf{T}) = \underbrace{\sigma(\text{DW}_{3\times3}(\mathbf{T}))}_{\text{Spatial}} + \underbrace{\sigma(\text{PW}(\text{AvgPool}(\mathbf{T}))) \odot \mathbf{T}}_{\text{Channel}}
\end{equation}
    Crucially, FARM computes the attention map using an \textit{additive similarity} function rather than matrix multiplication. The refined output $\mathbf{F}_{farm}$ is generated by modulating the Value branch with the aggregated context:
\begin{equation}
\mathbf{F}_{farm} = \text{Proj}(\Phi(\mathbf{Q}) + \Phi(\mathbf{K})) \odot \mathbf{V}
\end{equation}

    

\subsection{Decoupled Head}
Following the multi-scale feature fusion provided, the refined feature pyramid $\mathcal{P} = \{P_l\}_{l=3}^{5}$ is fed into the detection head. To resolve the conflict between the translation invariance required for classification and the translation variance required for localization, we adopt a decoupled head architecture. As illustrated in Figure~\ref{fig:head}, for each scale level $P_l$, the processing is split into two parallel streams:

\begin{figure}[h]
\centering
\includegraphics[width=0.5\textwidth]{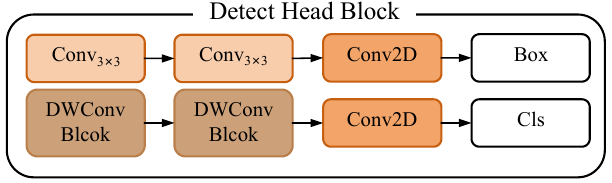}
\caption{The architecture of the Decoupled Detection Head, featuring parallel streams for classification and regression.} 
\label{fig:head}
\end{figure}

    $\mathcal{P} = \{P_3, P_4, P_5\} \in \mathbb{R}^{\frac{H}{2^l} \times \frac{W}{2^l} \times C_l}$ denote the input feature map at scale $l$. To accommodate the distinct requirements of the two tasks, the decoupled head processes the features through parallel streams with specialized architectural designs. Specifically, for the \textbf{Classification Branch}, we prioritize computational efficiency while preserving semantic discrimination by employing a Depth-wise Convolution ($\text{DWConv}$) followed by a $1\times1$ Convolution ($\text{Conv}_{2D}$). In contrast, the \textbf{Regression Branch} focuses on spatial precision for accurate bounding box estimation, utilizing a standard $3\times3$ Convolution ($\text{Conv}_{3\times3}$) followed by a $1\times1$ Convolution. Consequently, the resulting output tensors for classification, $\mathbf{O}_{cls}^l$, and regression, $\mathbf{O}_{reg}^l$, are formally obtained as:
\begin{align}
\mathbf{O}_{cls}^l &= \text{Conv}_{2D}(\text{DWConv}(\mathbf{X}_l)) \in \mathbb{R}^{H_l \times W_l \times C_{cls}} \\
\mathbf{O}_{reg}^l &= \text{Conv}_{2D}(\text{Conv}_{3\times3}(\mathbf{X}_l)) \in \mathbb{R}^{H_l \times W_l \times 4n}
\end{align}
where $C_{cls}$ denotes the number of defect categories, and $n$ represents the number of discretized bins used for the integral representation.



    The detection process operates simultaneously across the three granularities $\{P_3, P_4, P_5\}$. The outputs from all scales are decoded into a global set of candidate predictions $\mathcal{Y}_{raw} = \bigcup_{l=3}^{5} [\mathcal{Y}_l]$, where $\mathcal{Y}_l = (\mathbf{O}_{cls}^l, \mathbf{O}_{reg}^l)$.Each prediction consists of a bounding box, a class label, and a confidence score. To eliminate redundant detections of the same defect, we apply the \textbf{Non-Maximum Suppression (NMS)} mechanism. NMS iteratively selects the proposal with the highest confidence score and suppresses valid boxes that have an Intersection over Union (IoU) greater than a threshold $\theta_{nms}$ with the selected proposal. The final detection result $\mathcal{Y}_{final}$ is formulated as:
\begin{equation}
\mathcal{Y}_{final} = \text{NMS}(\mathcal{Y}_{raw}, \theta_{nms})
\end{equation}
This ensures that each micro-defect is represented by a single, most accurate bounding box.

\subsection{Optimization Objective}
    To train the CF-YOLO framework end-to-end, we employ a comprehensive multi-task loss function $\mathcal{L}_{total}$ that jointly optimizes both deterministic and probabilistic outputs. The total objective is defined as:
\begin{equation}
\mathcal{L}_{total} = \lambda_{box} \mathcal{L}_{CIoU} + \lambda_{cls} \mathcal{L}_{BCE} + \lambda_{dfl} \mathcal{L}_{DFL}
\end{equation}
where $\lambda_{box}$, $\lambda_{cls}$, and $\lambda_{dfl}$ are hyperparameters used to balance the contribution of each loss term. Specifically, the Regression Loss $\mathcal{L}_{CIoU}$ utilizes Complete-IoU to penalize the geometric misalignment between the predicted box $\hat{\mathbf{b}}$ and ground truth $\mathbf{b}_{gt}$. The Classification Loss $\mathcal{L}_{BCE}$ optimizes the category prediction probabilities using Binary Cross-Entropy. Finally, the Distribution Focal Loss $\mathcal{L}_{DFL}$ targets the Integral Representation in the regression branch, forcing the predicted distribution $\mathcal{P}(x)$ to sharpen around the target values. This enables the model to focus on refining the blurred boundaries typical of camouflaged defects.

\section{Experimental Results}\label{sec4}

\subsection{Dataset}
    Publicly available datasets for industrial copper tube defects are extremely scarce, which significantly hinders the development of data-driven detection algorithms. To address this gap and facilitate community research, we constructed and released the \textbf{Copper Tube Defect Dataset (CTDD)}. The dataset comprises a total of \textbf{1,847} high-resolution images and \textbf{4,898} single-class defect instances, captured under diverse industrial lighting conditions to simulate real-world manufacturing environments.

\begin{figure}[h]
    \centering
    \includegraphics[width=1\linewidth]{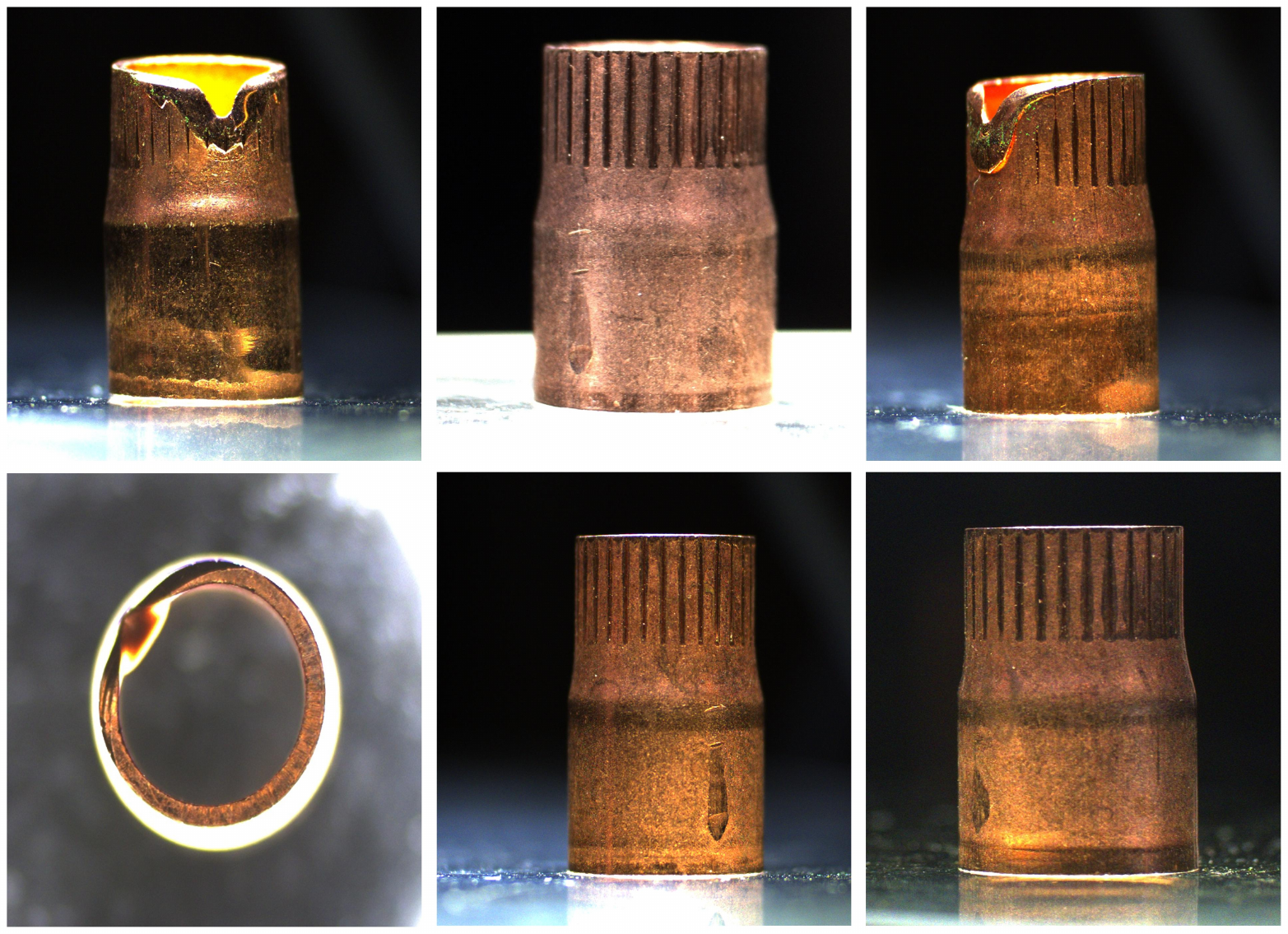}
    \caption{Visual samples from the constructed Copper Tube Defect Dataset.}
    \label{fig:Dataset}
\end{figure}

    To ensure annotation quality, all images were labeled by professional quality inspection engineers using the LabelImg tool, strictly adhering to the COCO annotation format. The dataset uses single-class defect annotation, with all annotated defect boxes assigned to one defect category. For experimental evaluation, we randomly partitioned the dataset into training, validation, and testing sets following a ratio of 8:1:1. This rigorous split ensures that the evaluation reflects the model's generalization capability on unseen data. Some representative visualization samples of the dataset are illustrated in Fig.~\ref{fig:Dataset}.
\subsection{Implementation Details}
    All experiments were conducted on a workstation equipped with an NVIDIA GeForce RTX 3090 GPU (24GB VRAM). The software environment consists of Python 3.10 and PyTorch 2.1.0 We implemented the proposed CF-YOLO and all baseline models based on the Ultralytics framework to ensure a fair comparison.

    During training, the input images were resized to $640 \times 640$ pixels. We employed the Stochastic Gradient Descent (SGD) optimizer with a momentum of 0.937 and a weight decay of 0.0005. The initial learning rate was set to 0.01 with a cosine annealing scheduler. All models were trained from scratch for 200 epochs with a batch size of 16. Standard data augmentation techniques, including Mosaic and Mixup\cite{bochkovskiy2020yolov4optimalspeedaccuracy}, were applied during the training phase to enhance model robustness, but were disabled during the final 10 epochs to ensure precise convergence.

\subsection{Evaluation Metrics}
    To comprehensively and rigorously assess the performance of the proposed method, we employ a multi-dimensional evaluation protocol comprising Precision, F1-score, Average Precision (AP) \cite{DBLP:journals/ijcv/EveringhamGWWZ10}, and Mean Intersection over Union (mIoU) \cite{DBLP:conf/eccv/LinMBHPRDZ14}.

    First, to evaluate the localization and classification capabilities, we adopt the standard \textbf{Mean Average Precision (mAP)}, specifically reporting \textbf{AP50}. This metric represents the average precision calculated at an Intersection over Union (IoU) threshold of 0.5. For a set of $N$ defect classes, the mAP is defined as the mean of the Average Precision (AP) for each class:
\begin{equation}
    \text{mAP} = \frac{1}{N} \sum_{i=1}^{N} \text{AP}_i @ \text{IoU}=0.5
\end{equation}
where $\text{AP}_i$ is the area under the Precision-Recall curve for the $i$-th class. AP50 is widely regarded as the \textit{de facto} standard in industrial defect detection for benchmarking model sensitivity. In addition, AP75 and AP95 denote AP evaluated at IoU thresholds of 0.75 and 0.95, respectively. 
    In industrial scenarios, controlling the false positive rate is as critical as ensuring high recall. Therefore, we utilize \textbf{Precision}, \textbf{F1-score}, and \textbf{F0.5-score} to evaluate the model's reliability. Precision measures the proportion of true positive predictions among all positive predictions, while the F1-score provides a harmonic mean of Precision and Recall. F0.5 gives greater weight to Precision, which is important in industrial inspection where false alarms can interrupt production:
\begin{gather}
    \text{Precision} = \frac{TP}{TP + FP} \\
    \text{F1} = 2 \cdot \frac{\text{Precision} \cdot \text{Recall}}{\text{Precision} + \text{Recall}} \\
    \text{F}_{0.5} = (1 + 0.5^2) \cdot \frac{\text{Precision} \cdot \text{Recall}}{0.5^2 \cdot \text{Precision} + \text{Recall}}
\end{gather}
where $TP$ and $FP$ denote True Positives and False Positives, respectively. A high F1-score indicates that the method achieves a robust trade-off between minimizing missed detections and reducing false alarms.

    Finally, to assess box-level localization consistency, we employ the \textbf{Mean Intersection over Union (mIoU)}. Since CTDD provides bounding-box annotations rather than segmentation masks, mIoU is computed between predicted bounding boxes and ground-truth bounding boxes:
\begin{equation}
\begin{aligned}
\mathrm{mIoU}
&=
\frac{1}{M}
\sum_{j=1}^{M}
\frac{\lvert \hat{B}_j \cap B_j \rvert}
     {\lvert \hat{B}_j \cup B_j \rvert}
\end{aligned}
\end{equation}
where $\hat{B}_j$ and $B_j$ denote the predicted and ground-truth bounding boxes for the $j$-th matched defect instance, respectively. This metric evaluates box-level spatial alignment rather than pixel-level mask overlap or defect-shape conformity.

\begin{table*}[h]
    \centering
    \caption{ {Performance comparison with evaluated detection methods on CTDD. The best result is highlighted in bold, and the second-best result is underlined.}}
    \label{tab:sota}
\resizebox{1\textwidth}{!}{ 
    \setlength{\belowcaptionskip}{10pt}
    \small
    \renewcommand{\arraystretch}{1.25}
    \begin{tabular}{lcccccccc}
        \toprule
        \textbf{Model} & \textbf{P} & \textbf{AP$_{50}$} & \textbf{AP$_{95}$} & \textbf{AP$_{75}$} & \textbf{mIoU} & \textbf{F$_1$} & \textbf{F$_{0.5}$} & \textbf{Avg} \\
        \midrule
        Cascade R-CNN~\cite{DBLP:conf/cvpr/CaiV18} & 0.594 & 0.732 & 0.479 & 0.334 & 0.755 & 0.661 & 0.619 & 0.596 \\
        Faster R-CNN~\cite{DBLP:conf/nips/RenHGS15} & 0.500 & 0.746 & 0.471 & 0.265 & 0.613 & 0.537 & 0.737 & 0.553 \\
        CenterNet~\cite{DBLP:conf/iccv/DuanBXQH019} & \underline{0.853} & 0.667 & 0.404 & 0.221 & 0.753 & 0.720 & 0.798 & 0.631 \\
        ATSS~\cite{DBLP:conf/cvpr/ZhangCYLL20} & 0.832 & 0.483 & 0.402 & 0.081 & 0.706 & 0.638 & 0.471 & 0.516 \\
        RetinaNet~\cite{DBLP:conf/iccv/LinGGHD17} & 0.448 & 0.684 & 0.445 & 0.152 & 0.704 & 0.568 & 0.489 & 0.499 \\
        Dynamic R-CNN~\cite{DBLP:conf/eccv/ZhangCMWC20} & 0.547 & 0.695 & 0.450 & 0.259 & 0.731 & 0.644 & 0.582 & 0.558 \\
        FoveaNet~\cite{DBLP:conf/iccv/LiJWLYSLCYF17} & 0.479 & 0.659 & 0.428 & 0.131 & 0.705 & 0.577 & 0.514 & 0.499 \\
        ETDNet~\cite{9353254} & 0.675 & 0.788 & \textbf{0.560} & 0.411 & 0.775 & 0.653 & 0.666 & 0.647 \\
        Conditional DETR~\cite{DBLP:conf/iccv/MengCFZLYS021} & 0.546 & 0.582 & 0.268 & 0.291 & 0.663 & 0.600 & 0.566 & 0.476 \\
        RF-DETR & 0.843 & 0.794 & 0.441 & 0.412 & \textbf{0.789} & \underline{0.790} & 0.778 & 0.692 \\
        RT-DETRv2 & 0.801 & 0.787 & 0.435 & 0.413 & \underline{0.782} & 0.770 & 0.766 & 0.679 \\
        YOLOv11n~\cite{khanam2024yolov11overviewkeyarchitectural} & 0.843 & \underline{0.801} & 0.464 & \underline{0.429} & 0.780 & 0.771 & \underline{0.806} & \underline{0.699} \\
        \textbf{CF-YOLO (Ours)} & \textbf{0.882} & \textbf{0.823} & \underline{0.488} & \textbf{0.455} & 0.769 & \textbf{0.796} & \textbf{0.846} & \textbf{0.723} \\
        \bottomrule
    \end{tabular}
    }
    \normalsize
    \vspace{1em}
\end{table*}
\subsection{Comparison with Evaluated Detection Methods}
     {To validate CF-YOLO in practical industrial scenarios, we conducted a comparative study against established detectors and recent representative methods under the same CTDD split where executable implementations were available.} The comparison includes:
\begin{itemize} 
    \item \textbf{Established CNN-based Baselines:} {We benchmark a comprehensive suite of seminal detectors to serve as robust baselines. This includes high-precision \textbf{two-stage architectures} (i.e., Faster R-CNN, Cascade R-CNN, Dynamic R-CNN) and representative \textbf{one-stage or anchor-free paradigms} (i.e., RetinaNet, ATSS, CenterNet, FoveaNet), covering a wide spectrum of dense prediction strategies.}
    \item \textbf{ {Recent architectures:}} { {To capture recent advancements in detection methodology, we incorporate YOLOv11n, RF-DETR, RT-DETRv2, Conditional DETR, and ETDNet. YOLOv11n is used as the primary baseline because CF-YOLO is implemented by modifying the YOLOv11 detection pipeline, which isolates the effect of CPAM and FARM within a closely matched detector family.}}
\end{itemize}

    To ensure a fair and rigorous comparison, all baseline methods were evaluated using their official hyperparameter configurations, while strictly sharing the identical data partition on the CTDD training and testing sets. The quantitative comparison results are reported in Tab.~\ref{tab:sota}.
    
     {As evidenced by the results, CF-YOLO obtains the best Precision, AP50, AP75, F1-score, F0.5-score, and Avg among the evaluated CTDD baselines, while ETDNet obtains the highest AP95 and RF-DETR obtains the highest mIoU. Compared with YOLOv11n, CF-YOLO improves AP50 from 0.801 to 0.823, F1-score from 0.771 to 0.796, and Precision from 0.843 to 0.882. These results indicate that the proposed context-perception and feature-refinement design is effective on CTDD.}
\begin{figure*}[h]
\centering
\includegraphics[width=1\textwidth]{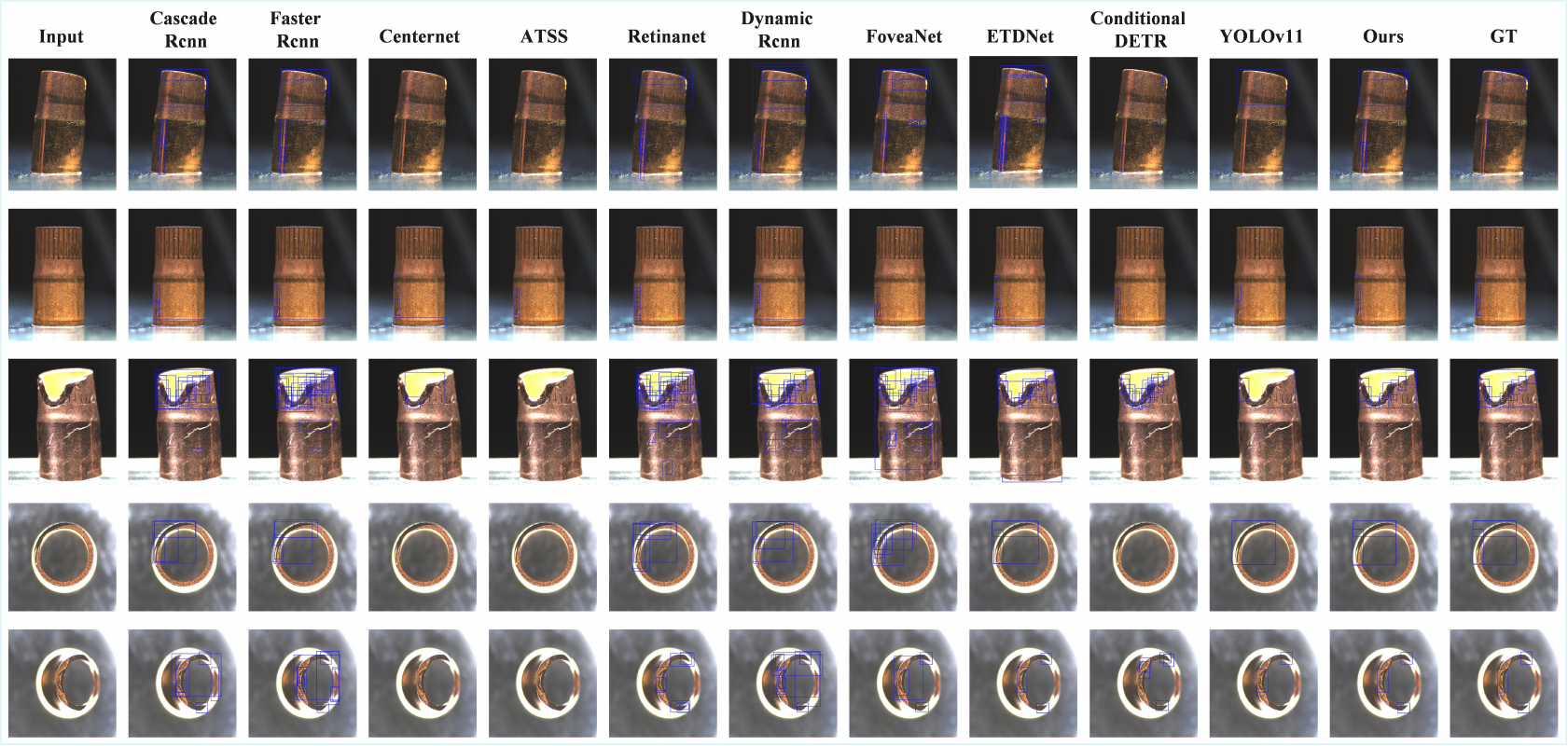}
\caption{Visualization of detection results compared against the Ground Truth (GT). The last column shows the GT annotations, while the second to last column shows the predictions of our proposed method. Preceding columns display results from other models for comparison} 
\label{main_result}
\end{figure*}
    To provide a more intuitive assessment of the proposed method's robustness in real-world industrial scenarios, we visualize the detection results of CF-YOLO alongside the baseline YOLOv11 and other evaluated models on the CTDD test set. As illustrated in Fig.~\ref{main_result}, the comparative results reveal distinct performance gaps under challenging conditions.
    
    It can be observed that baseline methods struggle significantly with micro-scale defects and low-contrast targets. In cases of background camouflage— {used here to describe high visual similarity between defects and the copper substrate rather than a generic COD benchmark setting}—mainstream detectors frequently suffer from missed and erroneous detections (False Negatives) due to insufficient feature extraction capabilities. Furthermore, they are prone to generating false positives in areas with complex surface oxidation or lighting noise. In contrast, thanks to the integrated attention mechanisms and multi-scale feature fusion, CF-YOLO exhibits strong sensitivity to subtle anomalies. Our method localizes many texture-similar defects and suppresses background interference under the evaluated CTDD protocol.

\subsection{External Validation on NEU-DET}
 {To provide an external check beyond CTDD, we evaluated CF-YOLO on the public NEU-DET industrial surface defect dataset~\cite{NEUDET}. Because the present work is formulated as single-class bounding-box defect detection, the six original NEU-DET defect categories were merged into a single \textit{defect} class. The resulting setting contains 1,800 images and 4,189 defect instances, split into 1,260 training images, 270 validation images, and 270 test images using seed 0.}

\begin{table}[htbp]
    \centering
    \caption{ {External validation results on NEU-DET.}}
    \label{tab:neu_det}
    \setlength{\tabcolsep}{4pt}
    \begin{tabular}{lccccc}
        \toprule
        \textbf{Model} & \textbf{AP$_{50}$} & \textbf{AP$_{50-95}$} & \textbf{F$_1$} & \textbf{P} & \textbf{R} \\
        \midrule
        YOLOv11n & 0.786 & 0.468 & 0.712 & 0.774 & 0.660 \\
        CF-YOLO & \textbf{0.794} & 0.465 & \textbf{0.732} & 0.760 & \textbf{0.707} \\
        \bottomrule
    \end{tabular}
\end{table}

 {
As shown in Tab.~\ref{tab:neu_det}, CF-YOLO improves AP50, F1-score, and Recall on NEU-DET, while AP50-95 and Precision are slightly lower than YOLOv11n. We therefore interpret this result as preliminary external validation rather than evidence of universal generalization across all industrial defect domains.
}

\subsection{Ablation of the Component}
\begin{figure}[h]
    \centering
    \includegraphics[width=1\linewidth]{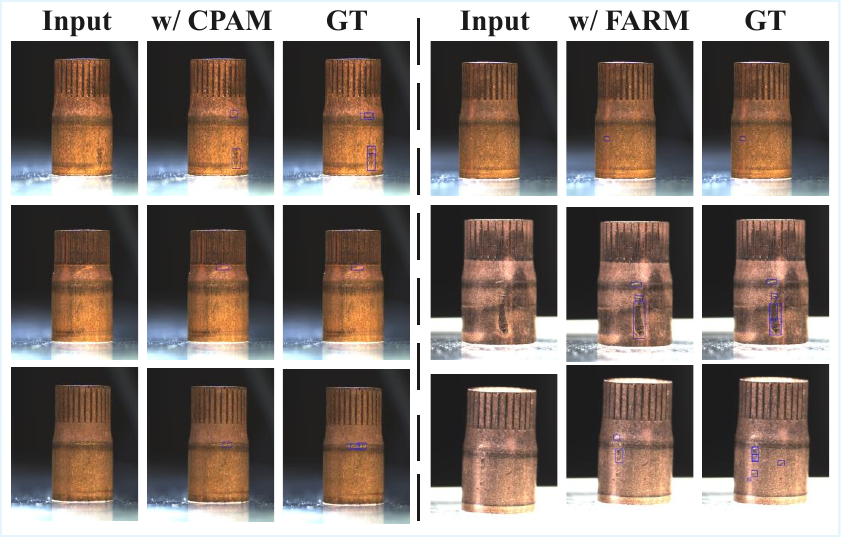}
    \caption{Visual ablation analysis of key components. The left panel demonstrates the effectiveness of CPAM in localizing camouflaged defects, while the right panel illustrates the capability of FARM in capturing micro-scale anomalies. 'GT' denotes Ground Truth.}
    \label{fig:component}
\end{figure}
    To rigorously verify the effectiveness of each proposed component within CF-YOLO, we conducted a comprehensive ablation study on the CTDD test set. We adopted the vanilla YOLOv11 as our strong baseline and incrementally incorporated the Context-Perception Aggregation Module (CPAM) and the Feature Additive Refinement Module (FARM). The quantitative comparison, focusing on Precision (P), F1-score, and AP@50, is summarized in Tab.~\ref{tab:ablation}.Additionally, qualitative visualizations verifying the specific contributions of CPAM and FARM are presented in Fig.~\ref{fig:component}.
    
    \textbf{Benefits of the CPAM:} As shown in the second row of Tab.~\ref{tab:ablation}, integrating CPAM into the backbone improves F1-score and AP@50 compared with the baseline. This gain is primarily attributed to the "Large-Kernel Perception" mechanism, which expands the receptive field to capture the macro-texture of the copper surface and helps preserve weak defect evidence.
    
    \textbf{Benefits of the FARM:} The third row demonstrates the impact of adding FARM alone. FARM improves F1-score and AP@50 in this component setting, indicating that additive feature refinement can help suppress false positives caused by intricate surface noise.
    
    \textbf{Synergistic Benefits:} Finally, the full CF-YOLO, which combines CPAM and FARM, achieves the highest performance across the reported component-ablation metrics. It surpasses the baseline by 3.9\%, 4.0\%, and 7.4\% in Precision, F1-score, and AP@50, respectively.

\begin{table}[htbp]
    \centering
    \caption{Ablation study on the effectiveness of proposed components.} 
    \setlength{\tabcolsep}{8pt} 
    \begin{tabular}{cc|ccc}
        \toprule
        \textbf{CPAM} & \textbf{FARM} & \textbf{P} &\textbf{F$_1$} & \textbf{AP$_{50}$}\\
        \midrule
        - & -   & 0.843 & 0.756 & 0.749\\
        $\checkmark$ & -  & 0.842 & 0.783 & 0.819\\
        -& $\checkmark$ & 0.838 & 0.780 & 0.800 \\
        $\checkmark$ & $\checkmark$ & \textbf{0.882} & \textbf{0.796} & \textbf{0.823}\\
        \bottomrule
    \end{tabular}
    \label{tab:ablation}
\end{table}

\subsection{Design Ablation Study}
    We further analyze the main design choices of CF-YOLO, including FARM branch composition, FARM placement, CPAM placement/quantity, and CPAM kernel configuration. The goal is to verify whether the adopted design is necessary rather than only reporting the final component combination.

    For FARM branch composition, we remove the spatial branch, remove the channel branch, and replace the dual-domain mixer with a standard convolution. As shown in Tab.~\ref{tab:farm_variants}, the complete FARM achieves the best performance across all metrics, while removing either branch degrades detection accuracy. This demonstrates the effectiveness of the dual-domain design for feature refinement.

\begin{table}[htbp]
    \centering
    \caption{FARM mixer variant ablation on CTDD. Full FARM is the unmodified configuration used in CF-YOLO.}
    \label{tab:farm_variants}
    \setlength{\tabcolsep}{4pt}
    \begin{tabular}{l c c c c c}
        \toprule
        \textbf{Variant} & \textbf{AP$_{50}$} & \textbf{AP$_{50\text{-}95}$} & \textbf{F$_1$} & \textbf{Precision} \\
        \midrule
        \textbf{Full FARM (Ours)}      & \textbf{0.823} & \textbf{0.488} & \textbf{0.796} & \textbf{0.882} \\
        No spatial branch & 0.772 & 0.467 & 0.748 & 0.825  \\
        No channel branch & 0.785 & 0.476 & 0.765 & 0.825 \\
        Conv-only       & 0.775 & 0.472 & 0.752 & 0.815 \\
        \bottomrule
    \end{tabular}
\end{table}

    We further evaluate three FARM placement strategies: before the neck, inside the neck, and after the neck. As shown in Tab.~\ref{tab:farm_placement}, placing FARM inside the neck yields the highest AP\textsubscript{50}, AP\textsubscript{50-95}, and F1-score. This indicates that feature refinement is most effective during multi-scale feature fusion.

\begin{table}[htbp]
    \centering
    \caption{FARM placement analysis on CTDD.}
    \label{tab:farm_placement}
    \setlength{\tabcolsep}{4pt}
    \begin{tabular}{l c c c c c}
        \toprule
        \textbf{Configuration} & \textbf{AP$_{50}$} & \textbf{AP$_{50\text{-}95}$} & \textbf{F$_1$}  \\
        \midrule
        \textbf{FARM in neck (Ours)} & \textbf{0.823} & \textbf{0.488} & \textbf{0.796}  \\
        FARM before neck & 0.760 & 0.469 & 0.738  \\
        FARM after neck  & 0.770 & 0.474 & 0.748  \\
        \bottomrule
    \end{tabular}
\end{table}

    For CPAM, we investigate different placement depths and block quantities. As reported in Tab.~\ref{tab:cpam_depth}, the mid-backbone configuration achieves the best performance, while reducing the number of CPAM blocks leads to a noticeable accuracy drop. These results confirm the importance of mid-level context perception and sufficient context aggregation.

\begin{table}[htbp]
    \centering
    \caption{CPAM depth and quantity analysis on CTDD.}
    \label{tab:cpam_depth}
    \setlength{\tabcolsep}{4pt}
    \begin{tabular}{l c c c c c}
        \toprule
        \textbf{Configuration} & \textbf{AP$_{50}$} & \textbf{AP$_{50\text{-}95}$} & \textbf{F$_1$} \\
        \midrule
        \textbf{CPAM mid-backbone (Ours)} & \textbf{0.823} & \textbf{0.488} & \textbf{0.796}  \\
        CPAM shallow  & 0.730 & 0.462 & 0.705  \\
        CPAM deep     & 0.718 & 0.460 & 0.695  \\
        Fewer CPAM blocks (2$\times$) & 0.770 & 0.468 & 0.745  \\
        \bottomrule
    \end{tabular}
\end{table}

    Finally, we evaluate different dual-kernel configurations of CPAM by varying the Small-Kernel Aggregation ($K_S$) and Large-Kernel Perception ($K_L$) sizes. As shown in Tab.~\ref{tab:ablation_kernel}, the adopted setting ($K_S=7$, $K_L=11$) achieves the best overall performance among all evaluated combinations. This suggests that effective defect detection relies on a balanced interaction between large-scale context perception and local feature aggregation, rather than larger kernels alone.



\begin{table}[htbp]
    \centering
    \caption{ {Ablation study on CPAM kernel settings. $K_L$ denotes the Large-Kernel Perception branch, and $K_S$ denotes the Small-Kernel Aggregation branch.}}
    \label{tab:ablation_kernel}
        \setlength{\tabcolsep}{4pt} 
        \begin{tabular}{cc|ccc}
            \toprule
            \textbf{$K_S$} & \textbf{$K_L$} & \textbf{P} & \textbf{F$_1$} & \textbf{AP$_{50}$}\\
            \midrule
             {3} &  {3}   &  {0.823} &  {0.703} &  {0.751} \\
            3  & 5  & 0.854 & 0.786 & 0.804 \\
            3  & 7  & 0.846 & 0.785 & 0.814 \\
            5  & 7  & 0.564 & 0.494 & 0.462 \\
            5  & 11 & 0.850 & 0.784 & 0.809 \\
             {11} &  {11} &  {0.844} &  {0.744} &  {0.783} \\
            \midrule
            \textbf{7} & \textbf{11} & \textbf{0.882} & \textbf{0.796} & \textbf{0.823} \\
            \bottomrule
        \end{tabular}
\end{table}
\section{Conclusions}
\label{sec5}

In this paper, we propose CF-YOLO, a specialized detector for industrial copper tube defect detection. By integrating the Context-Perception Aggregation Module (CPAM) and the Feature Additive Refinement Module (FARM), our method addresses the challenges of background camouflage and semantic ambiguity inherent in micro-defects. CPAM expands the receptive field to distinguish defects from the substrate, while FARM performs global semantic verification to suppress noise-induced false positives. Additionally, we released the Copper Tube Defect Dataset (CTDD) to alleviate data scarcity in this field. Extensive experiments show that CF-YOLO achieves the best AP50, Precision, and F1-score among the evaluated methods on CTDD while maintaining real-time inference speed on an RTX 3090.  {A limitation of CF-YOLO is that, while optimized for accurate real-time detection on GPU platforms, its deployment on resource-constrained edge devices may still be affected by the additional computational overhead of context aggregation and feature refinement. Future work will explore model compression and quantization-aware optimization to improve deployment efficiency.
}

\section*{Declarations}\label{Declarations}
\vspace{1\baselineskip}
{\bfseries Conflict of interest} The authors declare that they have no conflict of interest.\vspace{1\baselineskip}

\noindent{\bfseries Open Access} This article is licensed under a Creative Commons
Attribution 4.0 International License, which permits use, sharing, adaptation, distribution and reproduction in any medium or format, as
long as you give appropriate credit to the original author(s) and the
source, provide a link to the Creative Commons licence, and indicate if changes were made. The images or other third party material
in this article are included in the article’s Creative Commons licence,
unless indicated otherwise in a credit line to the material. If material
is not included in the article’s Creative Commons licence and your
intended use is not permitted by statutory regulation or exceeds the
permitted use, you will need to obtain permission directly from the copyright holder. To view a copy of this licence, visit \url{http://creativecomm ons.org/licenses/by/4.0/.}



\bibliography{ref}

\vspace{1\baselineskip}
\noindent{\bfseries Publisher's Note} Springer Nature remains neutral with regard to juris dictional claims in published maps and institutional affiliations.

\par\noindent 
\parbox[t]{\linewidth}{
\noindent\parpic{\includegraphics[width=1.5in,height=2in]{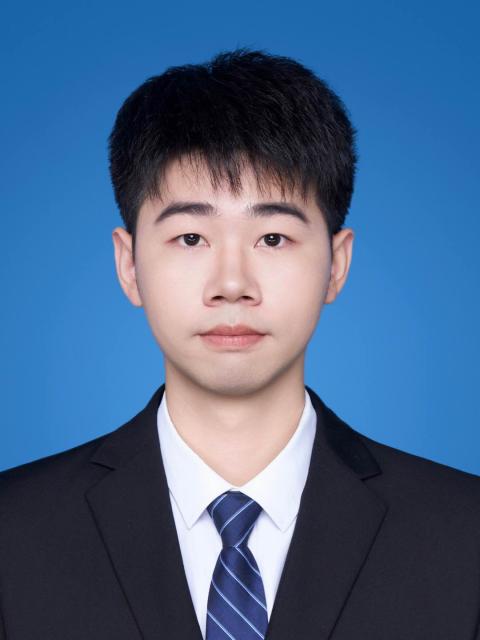}}
\noindent {\bf Xinda Yu}\
received his B.S. degree from Zhejiang Wanli University, China, in 2024. He is currently a masters student in HuZhou University. His research interests include computer vision, 3D reconstruction.}
\vspace{1\baselineskip}

\par\noindent 
\parbox[t]{\linewidth}{
\noindent\parpic{\includegraphics[width=1.5in,height=2in]{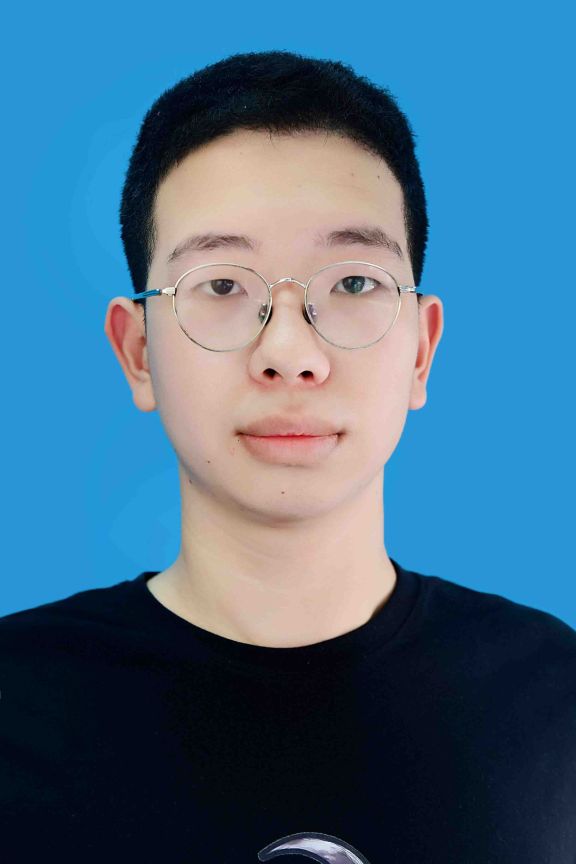}}
\noindent {\bf Kunxin Zheng}\
is currently a undergraduate student in HuZhou University. His research interests include computer vision.}
\vspace{6\baselineskip}

\par\noindent 
\parbox[t]{\linewidth}{
\noindent\parpic{\includegraphics[width=1.5in,height=2in]{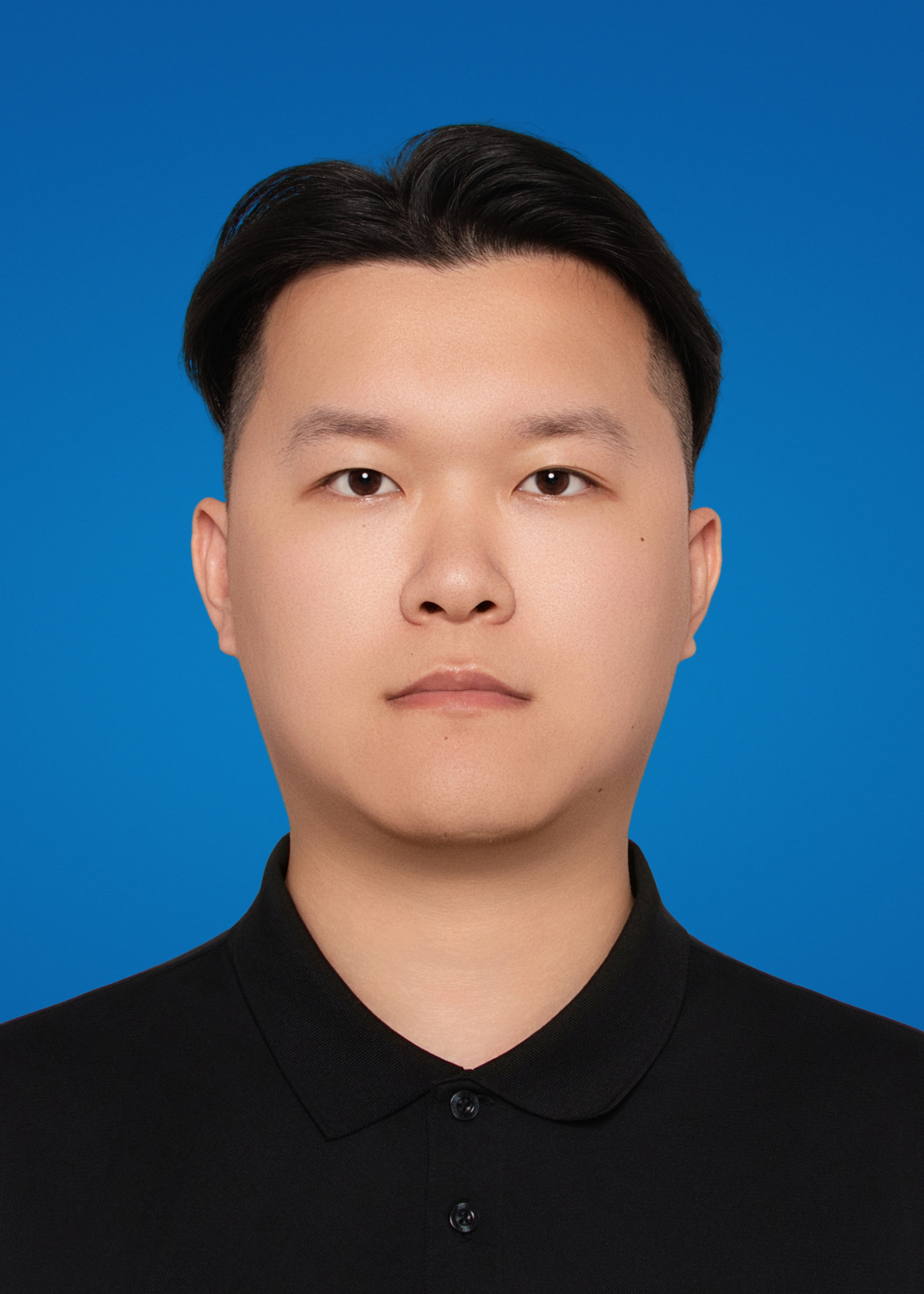}}
\noindent {\bf Chunan Yu}\
is currently pursuing the Ph.D. degree at Nanjing University of Science and Technology (NJUST). He received the M.S. degree of Huzhou University in 2025. He received B.S. degree from Linyi University in 2022. His research interests are in Computer Vision and Multimodal Learning.}
\vspace{1\baselineskip}

\par\noindent 
\parbox[t]{\linewidth}{
\noindent\parpic{\includegraphics[width=1.5in,height=2in]{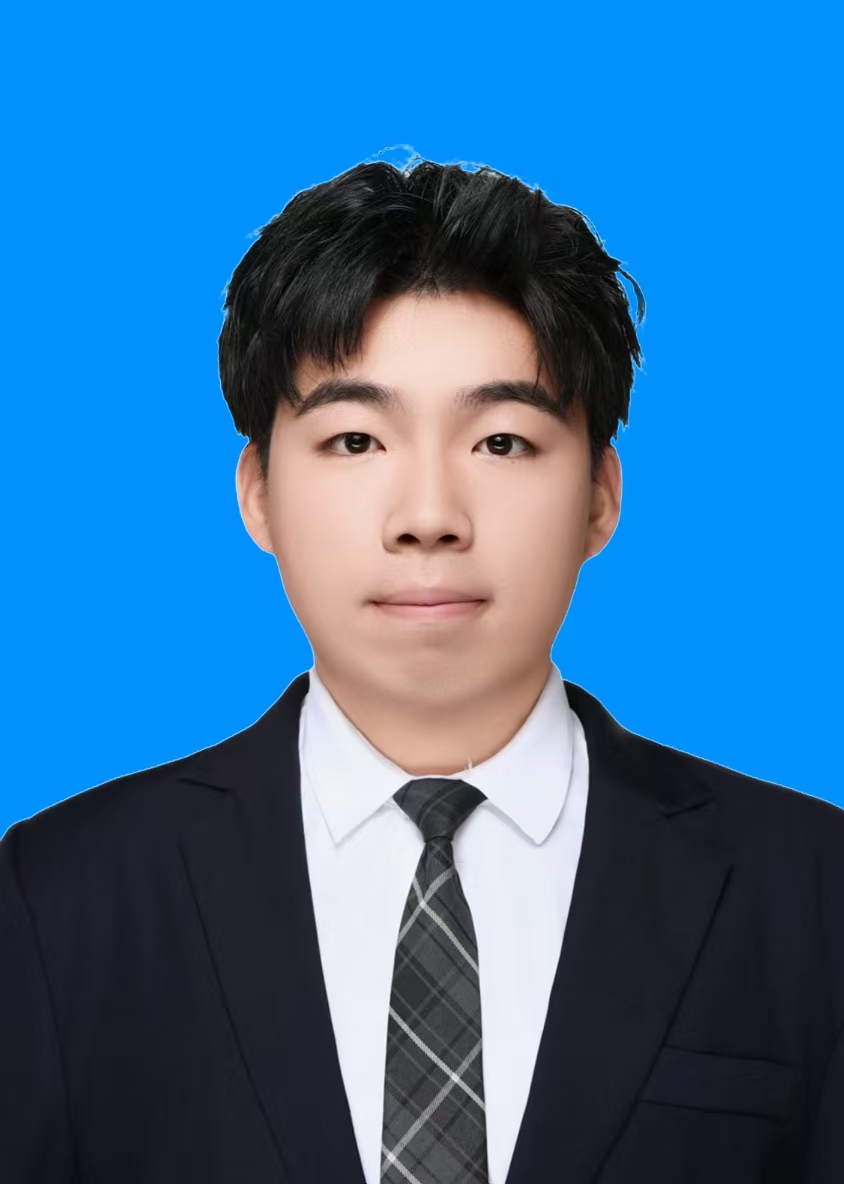}}
\noindent {\bf Qingbo Song}\
received his Bachelor of Engineering degree from Qilu Normal University, Jinan, China in 2024. He is currently a postgraduate student pursuing a master’s degree at Huzhou University, with research interests in computer vision and 3D reconstruction.}
\vspace{1\baselineskip}

\par\noindent 
\parbox[t]{\linewidth}{
\noindent\parpic{\includegraphics[width=1.5in,height=2in]{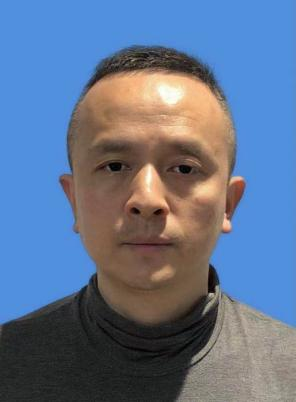}}
\noindent {\bf Hao Xiao}\
 received the M.S. degree in engineering from Wuhan University, China, in 2004; and the Ph.D. degree in management from Shanghai University of Finance and Economics, China, in 2014. He is currently an Associate Professor and Master's Supervisor. His research interests include software engineering, complex system modeling and analysis, and social computing.}
\vspace{1\baselineskip}

\par\noindent 
\parbox[t]{\linewidth}{
\noindent\parpic{\includegraphics[width=1.5in,height=2in]{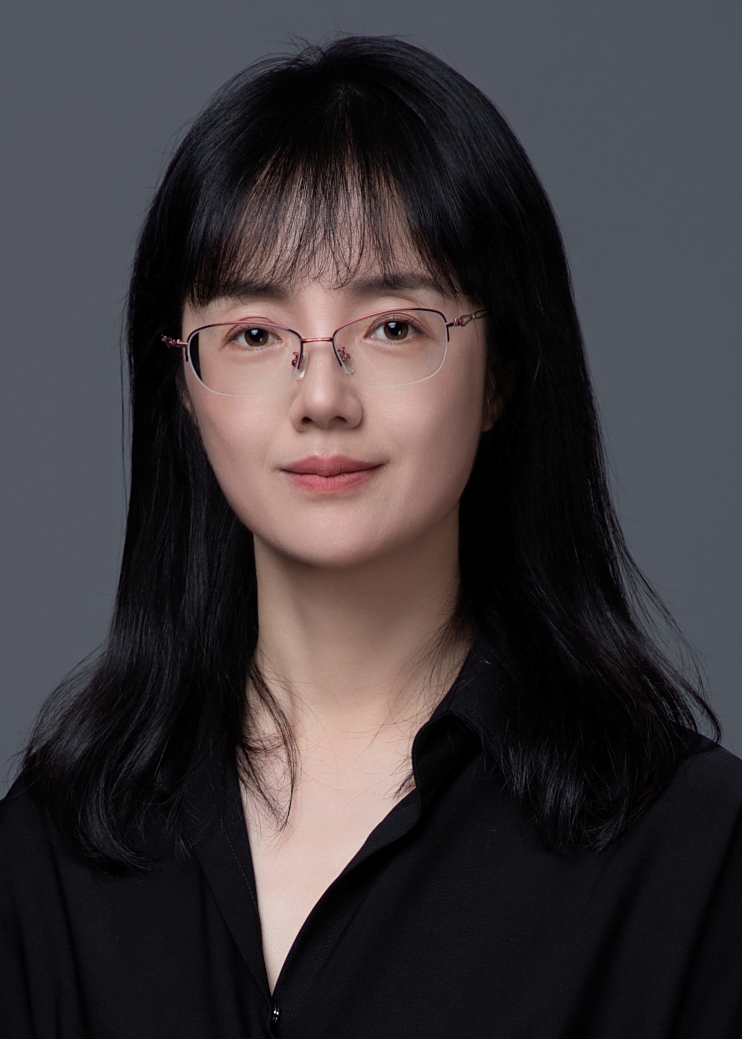}}
\noindent {\bf Ying Zang}\
received the B.E. degree in computer science and technology from Liaoning University, China, in 2005; the M.S. degree in computer science and technology from Dalian Maritime University, China, in 2010; and the Ph.D. degree in computer application technology from the University of Chinese Academy of Sciences, China, in 2022. She is currently a Lecturer and Master's Supervisor at the School of Information Engineering, Huzhou University. Her current research interests include object detection, semantic segmentation, and image processing.}
\vspace{1\baselineskip}

\par\noindent 
\parbox[t]{\linewidth}{
\noindent\parpic{\includegraphics[width=1.5in,height=2in]{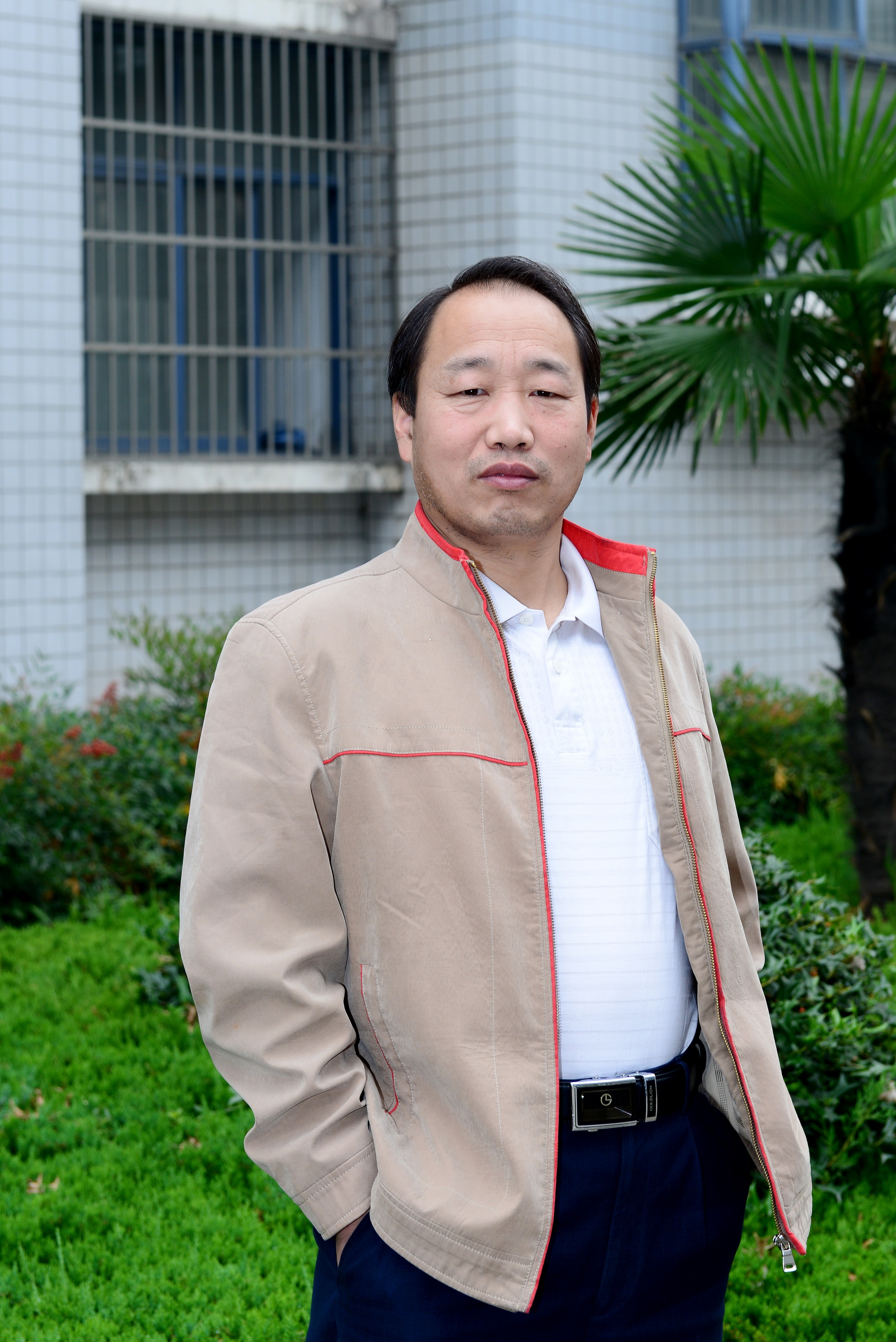}}
\noindent {\bf Jie Liu}\
received the B.S. degree in science from Fuyang Normal College, China, in 1993; and the M.S. and Ph.D. degrees in engineering from Hefei University of Technology, China, in 2004 and 2011, respectively. He is currently a Professor and Graduate Supervisor. He has presided over or participated in more than 10 national and provincial research projects, published over 50 papers, and holds 13 authorized invention patents. His current research interests include theoretical research and circuit design of RISC-V architecture processors and CNN accelerators, digital integrated circuit fault detection, and circuit design for intelligent instrumentation and industrial appliances.}
\vspace{1\baselineskip}

\end{document}